\documentclass{isprs} % isprs class modified 23-04-2019 (Dennis Wittich)
\usepackage{setspace}
\usepackage{geometry} % added 27-02-2014 Markus Englich
\usepackage{epstopdf}
\usepackage[labelsep=period]{caption}  % added 14-04-2016 Markus Englich - Recommendation by Sebastian Brocks
\usepackage[british]{babel} 
\usepackage[hang]{footmisc}
\usepackage[authoryear]{natbib}
\usepackage{textcase}

\usepackage[switch,pagewise]{lineno}

\usepackage{amsmath, bm}
\usepackage{mathbbol}
\usepackage{amssymb}
\usepackage{siunitx} % SI unit
\usepackage{svg}
\usepackage{float}
\usepackage{array}
\usepackage{subcaption}
\usepackage{verbatim} % comment
\usepackage[pdftex, english, pdfpagelabels=true]{hyperref}
\hypersetup{
    colorlinks = true
}
\usepackage{multirow}
\newcommand{\tildea}[1]{\overset{\sim}{#1}}
\newcommand{\tildeb}[1]{\stackrel{\sim}{\smash{#1}\rule{0pt}{1.1ex}}}

\usepackage{algorithm}
\usepackage{algpseudocode}
\algnewcommand\algorithmicforeach{\textbf{for each}}
\algdef{S}[FOR]{ForEach}[1]{\algorithmicforeach\ #1\ \algorithmicdo}
\DeclareMathOperator*{\argmax}{argmax} % thin space, limits underneath in displays

\begin{document}

%=========================
%==== Header =============

\title{LEVERAGING DYNAMIC OBJECTS FOR RELATIVE LOCALIZATION CORRECTION IN A CONNECTED AUTONOMOUS VEHICLE NETWORK}

\version{} % Leave empty (\version{}) when submitting the full paper

% KAO: Remove extra spacing
% Anonymous submissions, authors' names should not be visible
\author{ Y. Yuan\textsuperscript{1 \thanks{Corresponding author}} , M. Sester\textsuperscript{1}
}

% KAO: Remove extra newline
% Anonymous submissions, authors' affiliations should not be visible
\address{ \textsuperscript{1 }Institute of Cartography and Geoinformatics, Leibniz Universit\"at Hannover, Germany - (yuan, sester)@ikg.uni-hannover.de
}

% If the corresponding author is NOT the final author, always add a % space before the subsequent comma, i.e.
% first author name\textsuperscript{a,}\thanks{Corresponding author} , % second author name \textsuperscript{b}, etc.
% thanks to Niclas Borlin 05-05-2016

\commission{I, }{} %This field is optional. If filled, XX and YY should be replaced by adequate numbers. See https://www2.isprs.org/commissions/
\workinggroup{I/5} %This field is optional.
\icwg{}   %This field is optional.

\abstract{
% Background
High-accurate localization is crucial for the safety and reliability of autonomous driving, especially for the information fusion of collective perception that aims to further improve road safety by sharing information in a communication network of Connected Autonomous Vehicles (CAV). 
% Problem
In this scenario, small localization errors can impose additional difficulty on fusing the information from different CAVs.
% Our solution
In this paper, we propose a RANSAC-based (RANdom SAmple Consensus)  method to correct the relative localization errors between two CAVs in order to ease the information fusion among the CAVs. Different from previous LiDAR-based localization algorithms that only take the static environmental information into consideration, this method also leverages the dynamic objects for localization thanks to the real-time data sharing between CAVs. Specifically, in addition to the static objects like poles, fences, and facades, the object centers of the detected dynamic vehicles are also used as keypoints for the matching of two point sets. The experiments on the synthetic dataset COMAP show that the proposed method can greatly decrease the relative localization error between two CAVs to less than 20\SI{}{cm}  as far as there are enough vehicles and poles are correctly detected by both CAVs. Besides, our proposed method is also highly efficient in runtime and can be used in real-time scenarios of autonomous driving.
}


\maketitle

%\linenumbers
%=========================
%==== Content ============
\section{INTRODUCTION}\label{INTRO}
 
% KAO: Sloppy spacing ensures non-overfull lines. Can be removed if this is not an issue.
\sloppy
% importance of localization and current problems
Localization accuracy, robustness, and reliability are important criteria for the operation and the safety of autonomous vehicles, which need localization accuracy in the order of decimeters ($\sim$30cm) to stay in a specific lane \citep{MapBasedPV}. This is not an easy task.
% GNSS+IMU solution
The localization provided by the Global Navigation Satellite System (GNSS) can be easily influenced by weather and environmental situations such as buildings, trees, and tunnels. Its accuracy ranges from a few meters to over 20 meters \citep{GPSacc}. Solutions like Differential GPS (DGPS) can reduce the localization errors to less than two meters or more, but they need additional infrastructures and are not reliable all the time \citep{loc_survey}. IMU measures accelerations and angular velocities of the vehicle body which can be used to calculate the trajectory of the vehicle in a  Dead
Reckoning manner. However, it suffers from the error drifting effect. By integrating GNSS and IMU, \citep{gnss_imu1} reduced the localization error to several meters which is less than half of the result of the standalone solution with GNSS or IMU. Similarly, \citep{gnss_imu2} tested the localization errors of the integration system of GPS and a low-cost IMU (MPU6050). It shows an average error of 2.67\SI{}{m} and 0.96 \SI{}{\degree}. However, they tested in an open space where good GPS signals can be obtained. In urban areas, the performance of this solution can be worse when not enough satellites are in sight. To reduce the errors caused by GNSS signal interruptions, Tightly-Coupled (TC) fusion of GNSS and IMU uses filtering techniques to improve high-rate positioning results. The TC IMU/GNSS fusion framework proposed by \citep{gnss_imu3} can reduce the Root-mean-square error (RMSE) to less than 2\SI{}{m} when at least four satellites are visible and to 8\SI{}{m} when no satellite is available at a travel distance of 450\SI{}{m}. 

% GNSS+IMU+cameras/radar/etc.
To further improve the localization accuracy to the order of decimeters or even centimeters, more reference infrastructures such as base stations with know positions or onboard sensors such as cameras, Radars, or LiDARs are needed. 

On the one hand, Real-Time Kinematic (RTK) techniques improve the GNSS accuracy by the GNSS correction information received from GNSS base stations. With multiple GNSS receivers, it can reach an accuracy of centimeters. However, because of the limited communication range and power strength of base stations, it is hard or impossible to use this solution in some urban areas such as tunnels, tree canopies, or multi-path effect areas that are close to the buildings \citep{rtk1}. Besides, RTK-GNSS solutions also rely on the good visibility of GNSS signals. The accuracy decreases along with the outage duration of GNSS signals \citep{rtk2}.

% GNSS+IMU+radar and other sensors
On the other hand, methods that are only based on the onboard sensors either match the information of consecutive frames or register the local environment information to the global map to get the global localization. 
%  use Simultaneous Localization And Mapping (SLAM) technique to generate a local map and localize on it
Using the first method, \citep{radar} integrates short-range Radar with GPS/IMU and has reached an RMSE of 7.3\SI{}{m} and 37.7\SI{}{cm} laterally and longitudinally, respectively. But in the worst case, the longitudinal error has reached more than one meter. By registering the local radar reflection of underground features to the global underground feature map, a new method of using Localizing Ground-Penetrating Radar (LGPR) proposed by \citep{lgpr1} has shown a RMS error of 4\SI{}{cm} in different weather conditions and driving speed. However, the experiments in \citep{lgpr2} of using LGPR have reported a mean total error of 0.34\SI{}{m} in clear weather. In rainy weather, the result has degraded to 0.77\SI{}{m}. Different from LGPR and Radar, camera and LiDAR not only can be used for localization but also are necessary for the perception of the autonomous driving system. Therefore, it is convenient and efficient to use any stage of information from the perception process for localization. This information can both be used for local matching and global registration.

% GNSS+IMU+Camera
An early work by \citep{cam2009} has proposed a visual odometry method with a mono-camera to estimate the trajectory of vehicle motion. It estimates the relative locations of the same vehicle in consecutive frames by matching SIFT feature points with RANdom SAmple Consensus (RANSAC) algorithm. The corrected trajectory is visually improved. However, no error metric is reported in this work.
Instead of using RANSAC, \citep{cam2013} first matches GIST \citep{gist} descriptor representations of images to get coarse localization, which is then utilized as an initial localization for the fine localization step to find the nearest neighbors of SIFT \citep{sift} feature points. This work has achieved a localization error of 74\SI{}{cm}. 
To further improve the performance of image-based visual odometry, \citep{cam2017} integrates GPS and IMU with a camera sensor. It registers the detected lanes and symbolic road markings in the images to these in the digital High-Definition (HD) map. The average result of their experiments shows that the mean lateral error is about 0.5\SI{}{m} and the longitudinal error about 1.2\SI{}{m}. Furthermore, driven by deep-learning (DL) based object detection and semantic segmentation, more high semantic level of objects or points can be regarded as landmarks for the localization. \citep{QuadricSLAM} represents object-wise landmarks with 3D quadrics which are estimated by fusing the detected objects of images from different views. \citep{QuadricSLAM2} extended this work by decoupling the quadric parameters to improve the robustness against the observation noise. It reported an average ellipsoid central translation error of 2.13\SI{}{m}. 

In general, image-based visual odometry can reduce the localization error to less than one meter in some specific scenarios and can offer low-cost solutions. However, the accuracy of this method depends on the resolution of images. Higher resolution leads to higher accuracy but also to higher computational cost. Besides, small environmental changes such as illumination can introduce huge domain gaps to the captured images. This can greatly deteriorate the performance of the developed algorithms. In comparison to cameras, LiDARs actively sense the 3D environment by projecting 3D laser beams and collecting reflection information. The generated 3D point clouds are more accurate and more robust against weather changes than images. Therefore, they are more suitable for autonomous driving. Although they are currently high-cost sensors, the prices have a downward trend as many manufacturers are putting a big effort into decreasing the cost of LiDAR sensors.

Based on LiDAR sensors, \citep{lidar1} localizes vehicles by registering the curbs and road marking detected from point clouds data into the global feature map of the environment. It reduces the lateral and longitudinal errors to less than 30\SI{}{cm}. By the integration of GPS/IMU, filtering algorithms and more 3D features, the accuracy of LiDAR-based localization can be further improved. \citep{lidar2} uses 3D curb features, the intensity of road marker points as well as the height information of the surrounding high constructions to implement such an integration. It has achieved an accuracy of 9c\SI{}{cm} and 18\SI{}{cm} for lateral and longitudinal directions, respectively. \citep{lidar3} further improved the robustness of the localization against the dynamic changes of the environment by creating a probabilistic infrared intensity map for the road surface. Their result shows a positional RMSE of 9\SI{}{cm}. Similar to images, LiDAR-based localization also has many DL-based achievements \citep{lidar_dl1, lidar_dl2, lidar_dl3}, they all achieved better accuracy than the minimum required accuracy of 30\SI{}{cm} for autonomous driving. In summary, with better data, LiDAR-based methods can easily surpass the performance of image-based ones.

Both image- and LiDAR-based methods discussed previously suffer from the presence of dynamic objects such as cars, pedestrians, and the occlusions introduced by these objects. They attempt to reduce the negative influence of the dynamic objects either by predicting them and then removing them \citep{LoNet} or by selecting the most relevant features for localization\citep{DeepVCPAE}. Nevertheless, in a CAV network, these dynamic objects can be very useful information for localization because they are spatially better-distributed landmarks than most of the static features that might have unfavorable repetitive patterns for matching or registration, such as facades, curbs, and road markings.  

With CAV networks, or in other words, Vehicle-to-Vehicle (V2V) communications, some works \citep{v2v_loc1, v2v_loc2, v2v_loc3} have achieved the accuracy of several decimeters based on the travel properties of radio signals, such as time-of-arrival (TOA) and time-difference-of-arrival (TDOA).
% Knowledge gaps
However, this is still not accurate enough for the lane-keeping standard or for fusing the perception data coming from different vehicles that are supposed to improve the perception performance. Such collective perception can effectively reduce occlusion and increase road safety. However, \citep{v2vnet} show that data fusion among vehicles is very sensitive to localization errors. Especially, for object-wise fusion, a position error of only 0.4\SI{}{m} has led to a large performance drop in average precision of vehicle detection from about 85\% to 20\%. Therefore, high accurate localization is one of the key challenges for the perception system of autonomous vehicles to benefit from V2V communications. 

% Objective
In this paper, we propose an efficient method to correct the relative locations of vehicles in a CAV network. This correction can not only improve the information fusion performance of collective perception but also has the potential to improve the absolute localization by registering the fused data that has fewer occlusions, larger coverage, and more distinguishable patterns, to the global map. The proposed method uses the RANSAC algorithm to match the keypoints that are extracted from the result of the perception module of the autonomous system, specifically, the LiDAR-based object detection and semantic segmentation. It then uses the matched consensus to calculate the relative transformation between two vehicles.

% GNSS+IMU+LiDar
% V2I/V2V
% CP improves security but sensitive to localization errors, introduce proposed method

\section{METHOD}\label{sec:METHOD}

\subsection{Problem formulation}\label{sec:formulation}
In a mixed traffic scenario of collective perception which contains both autonomous vehicles and normal vehicles, we assume an ego CAV $C_0$ in the CAV networks. Other $N_\text{cav}$ CAVs ($C_i$, $i=1,..., N_\text{cav}$) that are in the communication range $R_\text{c}$ of the ego vehicle $C_0$ can share information to the ego vehicle.  The ground truth poses of these CAVs are notated as $X_i=(x_i, y_i, \theta_i)$, where $x_i$ and $y_i$ are the coordinates in the global coordinate system and $\theta_i$ is the orientation of the vehicle relative to the global x-axis.  Correspondingly, we use the symbol $\sim$ on the top of variables to represent the erroneous observations of these variables. Therefore, the observed erroneous poses by GPS/IMU system are noted as $\tildeb{X_i}=(\Tilde{x_i}, \Tilde{y_i}, \Tilde{\theta_i})$. 
All vehicles in the CAV network are collecting point cloud data which are notated as $PC_i$ ($i=0,..., N_\text{cav}$). Based on each point cloud $PC_i$, object detection, and point-wise semantic segmentation are implemented in the perception module of the autonomous system. From this module, we extract three kinds of point information for localization. First, the bounding box centers of the detected vehicles $P_\text{b}$, then the points of poles $P_\text{p}$ and the points of big planar structures $P_\text{f}$, such as facades and fences. Each point in the set $P_\text{b}$, $P_\text{p}$ and $P_\text{f}$ contains a vector of 2D $x$ and $y$ coordinates. We call $P_\text{b}$ and $P_\text{p}$ as anchor points for the simplicity of later explanation because they are the keypoints to be matched and then used for calculating the final transformation between the coordinate systems of the CAVs in the network.

The goal of the relative localization error correction is to find the optimized transformation parameters $\Delta x, \Delta y, \Delta \theta$ that transform the point set $P^{i,i}=\{P_\text{b}^{i,i}, P_\text{p}^{i,i}, P_\text{f}^{i,i}\}$ from the local coordinate system $CS_i$ of the cooperative vehicle to the coordinate system $CS_0$ of the ego vehicle so that the transformed point set $P^{i,0}$ has the maximum nearest neighborhood consensus with the point set of ego vehicle $P^{0,0}=\{P_\text{b}^{0,0}, P_\text{p}^{0,0}, P_\text{f}^{0,0}\}$. The superscripts of $P^{*,*}$ represents the index of vehicles and the coordinate systems. For example, $P^{i,0}$ is the points that are observed by $C_i$ and represented in $CS_0$.  The objective function of this optimization problem can be formulated as 

\begin{equation}
    Obj = \argmax_{\Delta x, \Delta y, \Delta \theta}\sum_{k=1}^{N_i} \mathbb{1}( \left\Vert \Delta T\cdot \tildeb{T}\cdot \tildeb{p_k}^{i,i}-\tildeb{p_k}^{0,0} \right\Vert^2 < \epsilon)
\end{equation}

\begin{equation}
    \tildeb{T} = \begin{bmatrix}
            cos(\Tilde{\theta_i} - \Tilde{\theta_0})) & -sin(\Tilde{\theta_i} - \Tilde{\theta_0}) & \Tilde{x_i} - \Tilde{x_0}\\
            sin(\Tilde{\theta_i} - \Tilde{\theta_0}) & cos(\Tilde{\theta_i} - \Tilde{\theta_0}) & 
            \Tilde{y_i} - \Tilde{y_0} \\
            0 & 0 & 1
        \end{bmatrix}
\end{equation}

\begin{equation}
    \Delta T = \begin{bmatrix}
            cos(\Delta \theta) & -sin(\Delta \theta) & \Delta x\\
            sin(\Delta \theta) & cos(\Delta \theta) & \Delta y \\
            0 & 0 & 1
        \end{bmatrix}
\end{equation}

where $\epsilon$ is the threshold for counting nearest neighbors and point $\tildeb{p_k}^{i,i}$ and $\tildeb{p_k}^{0,0}$ are the homogeneous representations of the original erroneous 2D points in set $\tildeb{P}^{i,i}$ and $\tildeb{P}^{0,0}$. $\tildeb{p_k}^{0,0}$ is the nearest neighbor of $\tildeb{p_k}^{i,0}=\Delta T\cdot \tildeb{T}\cdot \tildeb{p_k}^{i,i}$.

\subsection{Properties of relative errors}\label{sec:err_prop}
Since the true global location is not known, choosing the local coordinate system of any of the two CAVs as the reference for calculating the relative localization parameters will lead to an error propagation through a non-linear transformation from the global to the local coordinate system. The proof of this error propagation can be found in the Appendix.
Assume the localization error of both ego CAV $C_0$ and cooperative $C_i$ has the Gaussian distribution of $x,y\sim N(0, \sigma_{x,y}^2)\SI{}{m}$ and $\theta\sim N(0, \sigma_{\theta}^2)\SI{}{\degree}$. In order to fuse the point cloud data of the ego and cooperative CAV, the relative localization errors between the CAVs should be estimated. We accomplish this estimation in the coordinate system of ego vehicle $CS_0$, in which the three parameters  $\Delta x, \Delta y, \Delta \theta$ should be estimated so that $PC_i$ can be correctly transformed to $CS_0$ and better match $PC_0$. The derivation of the error propagation from the global to the local coordinate system can be found in the appendix A.

\begin{figure}[t]
\begin{center}
		\includegraphics[width=1.0\columnwidth]{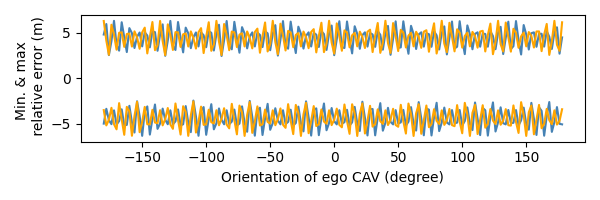}
	\caption{Minimum and Maximum relative localization error of two vehicles in x (blue) and y (orange) axis. The initial global localization errors $(-0.5\SI{}{m}, -0.5\SI{}{m}, -5\SI{}{\degree})$ and $(0.5\SI{}{m}, 0.5\SI{}{m}, 5\SI{}{\degree})$ are set for the two vehicles, respectively, as an example.}
\label{fig:rel_err_ex}
\end{center}
\end{figure}

Taking the error of $(-0.5\SI{}{m}, -0.5\SI{}{m}, -5\SI{}{\degree})$ for $C_0$ and $(0.5\SI{}{m}, 0.5\SI{}{m}, 5\SI{}{\degree})$ for $C_1$ as an example, the relative errors along x and y axis are shown in figure \ref{fig:rel_err_ex}, in which the relative translation error $\Delta x, \Delta y$ is in a range of about -6\SI{}{m} to 6\SI{}{m}. The magnitude of $\Delta x, \Delta y$ is positive correlated to the distance between $C_0$ and $C_i$ and is enlarged by this distance in comparison to the original error of 0.5\SI{}{m}. Besides, the translation errors oscillate as the orientation of the ego vehicle changes. Different to the translation errors, the rotation error $\Delta \theta$ keeps constant and equals to the summation of the absolute global localization error of both $C_0$ and $C_i$.

As a summary, the optimum relative errors $\Delta x, \Delta y, \Delta \theta$ should be searched in a large 3D space. It is time-consuming to conduct a refined search using a method like maximum consensus \citep{jedrick}, although it can provide the global optimum if the maximum search space is previously known.
For the example shown in figure \ref{fig:rel_err_ex}, the search space is about $12\SI{}{m}\times12\SI{}{m}\times10\SI{}{\degree}$. If the searching resolution is set to 0.1\SI{}{m} and $0.1\SI{}{\degree}$, the result of consensus should be counted for 1.44 ($120\times120\times100$) million times. Therefore, a more efficient method using RANSAC is proposed in this paper.

\subsection{Error correction with RANSAC}\label{sec:err_prop}

To obtain the classified points $P_\text{b}$, $P_\text{p}$ and $P_\text{f}$ for localization error correction, we use the pre-trained network that is proposed in \citep{keypoints_fusion} to generate bounding box predictions and the down-sampled points of poles, facades and fences. We use all the detected bounding boxes to generate $P_\text{b}$ and all points of poles for $P_\text{p}$. However, only a subset of $N_f$ points of facades and fences are selected using Furthest Point Sampling (FPS) for $P_\text{f}$ because these points can be any point on the planes and provide less accurate matches than $P_b$ and $P_p$, in which each point corresponds to the center of a specific object. A reduction in the number of points $P_f$ not only reduces the computational overhead but also makes the algorithm rely more on $P_b$ and $P_p$ so that it is more robust against the variation of point density in the point cloud data.

The workflow of proposed relative localization error estimation is described in algorithm \ref{alg:ransac_err_esti}. Line 1 and 2 define the notations of the algorithm. $\mathcal{A}$ and $\mathcal{B}$ are the set of the anchor points (center points of the detected vehicles and pole points) that belong to the ego point cloud $PC_0$ and the cooperative point cloud $PC_i$, respectively.  Line 3 defines the initial values that need to be updated during the workflow. For each point, $b$ in set $\mathcal{B}$, at most $n$ nearest neighbors are found in set $\mathcal{A}$ with a maximum Euclidian distance of $\epsilon_1$, because the correct matching points might not have the smallest distance among the $n$ nearest neighbors.We set $n$ to two to ensure the efficiency of the algorithm. Besides, the neighbors should have the same category label as point $b$ as described in line 5. The two neighbors $(\alpha, \beta)$ of $b$ are then put into set $M$ as dictionary format as shown in line 6. $\alpha$ and  $\beta$ can be None if not enough neighbors are found.

Based on the neighborhood matchings in set $M$,  we start our RANSAC matching iterations from line \ref{alg:itr_start}.
The total number of iterations is defined by the minimum value between the predefined maximum total number of iterations $N_\text{ransac}$ and the number of the combinations $C^2_{|M^\prime|}$ (line 9), where $|M^\prime|$ is the cardinality of $M^\prime$ (line 8).  

\begin{algorithm}[!t]
\caption{Relative localization error estimation}\label{alg:ransac_err_esti}
\begin{algorithmic}[1]
\Ensure \\ $\mathcal{A}=\tildeb{P_b}^{0,0}\cup \tildeb{P_p}^{0,0}$, 
$\mathcal{B}=\tildeb{P_b}^{i,0}\cup \tildeb{P_p}^{i,0}$,\\
$c_*$ : category label (box, pole or fence/facade) of point $*$, \\
$N_\text{cons,max}=0$, $M=\emptyset$.

\ForEach {$ b \in \mathcal{B}$}
    \State $(\alpha, \beta) \leftarrow NearestNeighbors(\mathcal{A}, b, 2)$, with
    $\left\Vert b-\alpha\right\Vert^2<\epsilon_1, 
    \quad \left\Vert b-\beta\right\Vert^2<\epsilon_1, \quad
    c_b=c_{\alpha}=c_{\beta}$
    \State $M\leftarrow \{b:(\alpha, \beta)\}\cup M$ \label{alg:M1}
\EndFor   
\State $M^\prime = \{b:a \mid a \in (\alpha, \beta) , b:(\alpha, \beta) \in M\}$
\State $N^\prime_\text{ransac}=min(N_\text{ransac},~C^2_{|M^\prime|})$
\ForAll {$i=1, \dots, N^\prime_\text{ransac}$}\label{alg:itr_start}
    \State $i1, i2 = Choice(M, 2)$
    \State $m_{i1}={b_{i1}:(\alpha_{i1},\beta_{i1})},\quad m_{i2}={b_{i2}:(\alpha_{i2},\beta_{i2})} $
    \State $a_{i1}=Choice((\alpha_{i1},\beta_{i1}), 1)$
    \State $a_{i2}=Choice((\alpha_{i2},\beta_{i2}), 1)$
    \State $T\leftarrow CalTF((a_{i1}, b_{i1}), (a_{i2}, b_{i2}))$
    \State $\mathcal{B}^\prime \leftarrow T \cdot (\mathcal{B}~\cup \tildeb{P_f}^{i,0}),
            \mathcal{A}^\prime \leftarrow \mathcal{A}~\cup \tildeb{P_f}^{0,0}$
    \State $N_\text{cons}=\sum_{k=1}^{|\mathcal{B^\prime}|} \mathbb{1}(\left\Vert b_k - a_k\right\Vert <\epsilon_2)$
    \If{$N_\text{cons, max}<N_\text{cons}$}
    \State $M^\prime=\{b:a \mid \left\Vert b - a \right\Vert^2 < \epsilon_2, a\in\mathcal{A}, b\in \mathcal{B}\}$
        \If{$|M^\prime|\leq 2$}
            \State $T_\text{out}\leftarrow CalTF(\{(a, b)\mid b:a \in M^\prime\})$
        \Else
            \State $T_{out}=I_3$
        \EndIf
    \EndIf
\EndFor

\State\Return $T_\text{out}$
\label{ag:return}
\end{algorithmic}
\end{algorithm}

In each iteration, two indices of the matches in set $M$ are randomly selected (line 11). The resulting matches chosen are given in line 12. Since the two nearest neighboring points $(\alpha, \beta)$ are found for each point $b$, only one from these two is selected for the current iteration (line 13-14) to calculate the relative transformation between the set $\mathcal{A}$ and $\mathcal{B}$ (line 15). Afterwards, the full selected localization points $\tildeb{P}^{i,0}=\{\tildeb{P_b}^{i,0}, \tildeb{P_p}^{i,0}, \tildeb{P_f}^{i,0}\}=\mathcal{B}~\cup \tildeb{P_f}^{i,0}$ are transformed with the resulting transformation matrix $T$ (line 16). The outcome $\mathcal{B}^\prime$ as well as the full localization point set of ego vehicle $\mathcal{A}^\prime=\tildeb{P}^{0,0}=\{\tildeb{P_b}^{0,0}, \tildeb{P_p}^{0,0}, \tildeb{P_f}^{0,0}\}=\mathcal{A}~\cup \tildeb{P_f}^{0,0}$ are used for counting the consensus for this iteration of RANSAC matching (line 17), for each point $b_k$ in $\mathcal{B}^\prime$, if it has a nearest neighbor $a_k$ whose distance to $b_k$ is less than the threshold $\epsilon_2$, the counting variable $N_\text{cons}$ will be increased by 1. If the counting result $N_\text{cons}$ is larger than the last iteration, the matching correspondences $M^\prime$ of set $\mathcal{A}$ and $\mathcal{B}$  and the refined transformation matrix $T_\text{out}$ should be updated (line 18-25). Different to $T$ which is calculated only based on two matched point pairs, $T_\text{out}$ uses all consensus anchor point pairs found between $\mathcal{A}$ and $\mathcal{B}$ to calculate a refined transformation. There are three parameters that need to be determined by $CalTF$ with at least two point pairs. Therefore, it is possible that not enough matchings are found for the refined error correction. In this case, we return an identity matrix as shown in line 21.

The detailed implementation of function $CalTF$ is described in algorithm \ref{alg:est_tf}. The input of the function is the $N$ matched 2D point pairs. The mean ($\overline{a}, \overline{b}$) and the reduced coordinates ($a_i^\prime, b_i^\prime$) of there points are first calculated respectively (lines 2-3). Then the correlations between x- and y-coordinates of the reduced points are calculated (lines 4-7). The rotation between the point set $A=\{a_i \mid i=1,\dots, N\}$ and $B=\{b_i \mid i=1,\dots, N\}$ can then be calculated with the equation described in line 8. Finally, the equation in line 9 calculates the translation between the two point sets. The returned transformation matrix is composed from the parameter $\Delta x, \Delta y$ and $\Delta\theta$.

\begin{algorithm}[!h]
\caption{Similarity transformation}\label{alg:est_tf}
    \begin{algorithmic}[1]
    \Procedure{CalTF}{Point pairs $M=\{(a_i,b_i) \mid i=1,\dots, N\}$}
    \State $\overline{a} = \frac{\sum_{i=1}^N a_i}{N}$, \quad $\overline{b} = \frac{\sum_{i=1}^N b_i}{N}$
    \State $a_i^\prime = a_i - \overline{a}$,\quad $b_i^\prime = b_i - \overline{b}$
    \State $S_{xx} = \sum_{i=1}^N (a_{i,x}^\prime \cdot b_{i,x}^\prime)$
    \State $S_{yy} = \sum_{i=1}^N (a_{i,y}^\prime \cdot b_{i,y}^\prime)$
    \State $S_{xy} = \sum_{i=1}^N (a_{i,x}^\prime \cdot b_{i,y}^\prime)$
    \State $S_{yx} = \sum_{i=1}^N (a_{i,y}^\prime \cdot b_{i,x}^\prime)$
    \State $\Delta\theta = arctan2(S_{xy}-S_{yx}, S_{xx}+S_{yy})$
    \State $[\Delta x, \Delta y] = \overline{b} - 
            \begin{bmatrix}
            cos\Delta\theta & -sin\Delta\theta \\
            sin\Delta\theta & cos\Delta\theta
            \end{bmatrix} \cdot \overline{a}$
    
    \State \Return $T=
            \begin{bmatrix}
            cos\Delta\theta & -sin\Delta\theta & \Delta x\\
            sin\Delta\theta & cos\Delta\theta & \Delta y \\
            0 & 0 & 1\end{bmatrix}$
    \EndProcedure
    \label{ag:return}
    \end{algorithmic} 
\end{algorithm}

\section{Experiments}\label{sec:experiments}
\subsection{Dataset}
We use a synthetic dataset COMAP \citep{comap} that is specially designed for collective perception with CAVs to validate the effectiveness of our proposed method. The reasons for this are two folds: first, it is extremely expensive to collect real data with networked CAVs; second, the synthetic dataset can also provide 100\% accurate ground truth localization which is beneficial for the evaluation of the localization error estimation algorithm.

COMAP contains 7788 frames of data collected by 20 CAVs and from 21 different junctions scenarios. Since 4155 frames are used for training the network proposed as in \citep{keypoints_fusion}, we use the other 2148 frames from the test set to evaluate our method. In each frame, we select one ego vehicle and at most five cooperative vehicles that are in the communication range $R_\text{c}=40\SI{}{m}$ of the ego vehicle for evaluation. The relative localization error between each pair of ego and the cooperative vehicle is estimated using algorithm \ref{alg:ransac_err_esti}.

% result figure
\begin{figure*}[!t]
\begin{center}
    \includegraphics[trim=1.5in 0.2in 1.5in 0.5in, clip=true,width=1.0\textwidth, height=4.5cm]{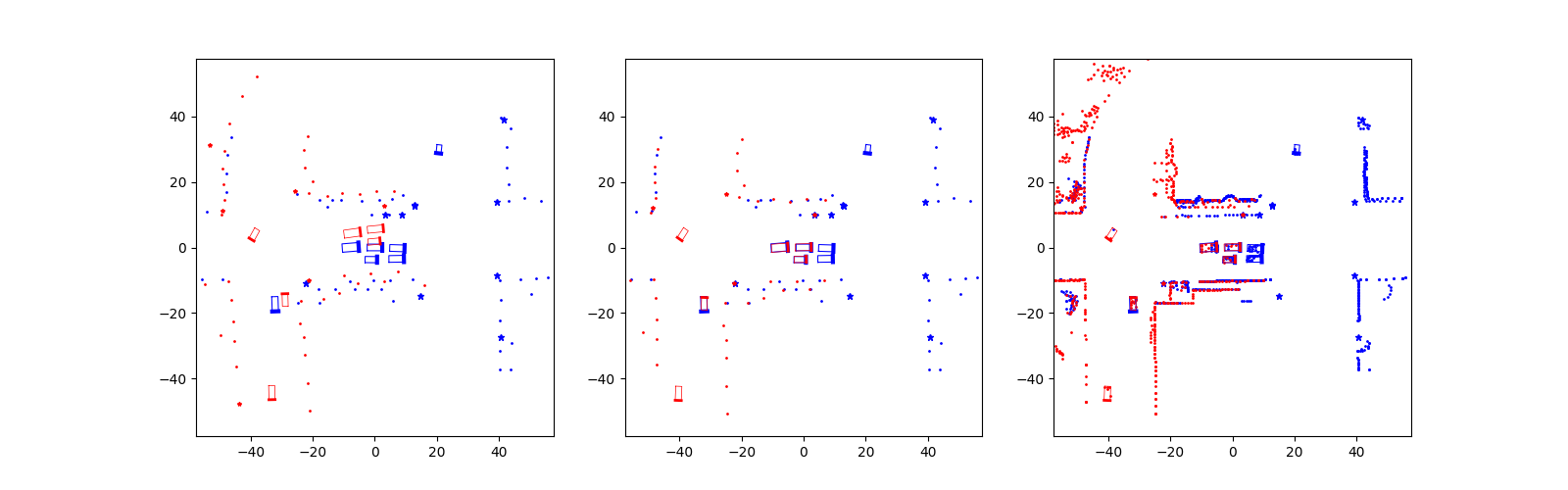}
	\caption{Sample result before (left) and after (middle and right) the relative localization error correction}
\label{fig:sample_result}
\end{center}
\end{figure*}
%result figure

\subsection{Experiment settings}
To generate efficient number of labeled points for localization, we empirically set $N_f=50$ which ensures that algorithm \ref{alg:ransac_err_esti} finds the correct consensus for $P_f$ with the lowest number of points. For nearest neighbor searching, the distance threshold $\epsilon_1$ is set according to equation
\begin{equation}\label{eq:eps1}
    \epsilon_1 = \eta \cdot R_\text{c} \cdot \sigma_r \cdot \pi / 180\SI{}{\degree}
\end{equation}
where $R_\text{c}$ is the communication range and $\sigma_r$ the standard deviation of the relative rotation error in degree. $R_\text{c} \cdot \sigma_r \cdot \pi/ 180\SI{}{\degree}$ is the standard deviation of the relative location error derived from $\sigma_r$. The factor $\eta=2.58$ is the z-score of standard normal distribution at the confidence level of 99\%,which means that the relative positin error caused by $\sigma_r$ has a confidence of 99\% to be less than  $\epsilon_1$. This setting ensures that RANSAC has a high chance to find correct point matchings in a limited searching range. After the rough error correction with the estimated transformation matrix $T$, the threshold $\epsilon_2$ for the second round of nearest neighbor searching is reduced to $1m$. 
To investigate the influence of the number of RANSAC iterations ($N_{ransac}$) and different standard deviations for localization errors ($\sigma_x, \sigma_y, \sigma_r$) on the performance of the proposed algorithm, we configure our experiments by varying $N_{ransac}$ from 10 to 50 with an offset of 10, 
and assuming the localization errors are normally distributed and varying
the standard deviation of the translation error $\sigma_x$ and $\sigma_y$ from 0.2\SI{}{m} to 1.0\SI{}{m} with an offset of 0.2\SI{}{m}, and the rotation error $\sigma_r$ from 2$\SI{}{\degree}$ to 10$\SI{}{\degree}$ with an offset of 2$\SI{}{\degree}$. All experiment runs are processed on a 6-Core Intel i7-8700 CPU.

\subsection{Evaluation metric}
The Root Mean Squared Error (RMSE) is used to evaluate localization accuracy after the error correction with the proposed algorithm. In order to have an intuitive comparison with the originally introduced localization errors in the global coordinate system, we take the residuals between the estimated and the ground truth relative errors that are represented in the global coordinate system for calculating RMSE.

However, the randomness of RANSAC might lead to failures in finding the optimal result. Besides, it is also possible that the field-of-view (FOV) of the two vehicles do not have sufficient overlapping detected vehicles and poles to calculate the relative localization error between them. Therefore, we also evaluate the rate of valid error correction results by filtering the results with the consensus numbers $N_\text{cons}$. Specifically, the valid rate is calculated with
\begin{equation}
    \gamma = \frac{\sum_1^N \mathbb{1}(N_\text{cons} > thr_\text{cons})}{N_\text{s}}
\end{equation}

where $N_\text{s}$ here is the total number of sample pairs of ego and cooperative vehicles, between which the relative localization error should be corrected, and $thr_\text{cons}$ is the threshold to filter the valid sample pairs. Finally, we use the processing frames per second (FPS) to evaluate the runtime of the proposed method.

\section{RESULTS AND EVALUATION}\label{sec:results}
In general, the proposed method can correct the relative errors to less than 20\SI{}{cm} (see figure \ref{fig:N_ransac}) as far as there are enough localization anchor points of the ego and cooperative vehicles can be matched and these matches have a good spatial distribution. An example result is shown in figure \ref{fig:sample_result}. All observed information of the ego vehicle is shown in blue, and cooperative vehicle in red. The detected vehicles are notated with rotated bounding boxes (BBoxes). The thick bars of the BBoxes indicate the front of the vehicle. Poles are shown with stars and other small points in the left two figures are the down-sampled points of fences and facades. Besides, all points before down-sampling are shown in the right subplot in order to better illustrate the performance of the error correction algorithm. As shown in the left subplot, the large initial relative error has led to wrong matches between the detected vehicles of the ego and the cooperative CAVs. After error correction, they are well matched as shown in the middle subplot. The right subplot shows that the point clouds of the two CAVs are also well aligned to each other.

To further give some quantitative details about the result, in the following subsections, we first show the relative error properties through error propagation based on the dataset. Then the result of parameter studies of the proposed method is demonstrated.

\subsection{Error propagation}
\begin{figure}[b]
\begin{center}
		\includegraphics[width=0.8\columnwidth, height=3.5cm]{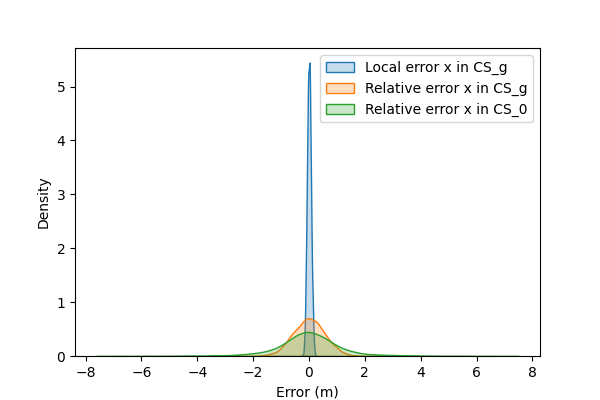}
	\caption{PDFs of errors in x-axis with initial local error of 0.4m}
\label{fig:rel_err}
\end{center}
\end{figure}
As discussed in section \ref{sec:err_prop}, the relative localization error between two vehicles can be amplified along with the error propagation through the non-linear transformation. This is also verified with the test dataset. After adding the normally distributed error with $\sigma_{x,y}=0.4m$ and $\sigma_r=4\SI{}{\degree}$ to the localization of each vehicle, the standard deviation of relative localization errors is greatly increased as shown in figure \ref{fig:rel_err}. The blue high peak shows that most vehicles have a very small initial localization error in the global coordinate system ($CS_g$), the addition of the errors of two vehicles is much more flatly distributed with a bigger standard deviation (orange). After transforming this relative error from the $CS_g$ to the local coordinate system $CS_0$, the standard deviation of the errors increased again (green).

\subsection{Performance on different RANSAC iterations}

\begin{figure}[th]
\begin{center}
		\includegraphics[width=1.0\columnwidth]{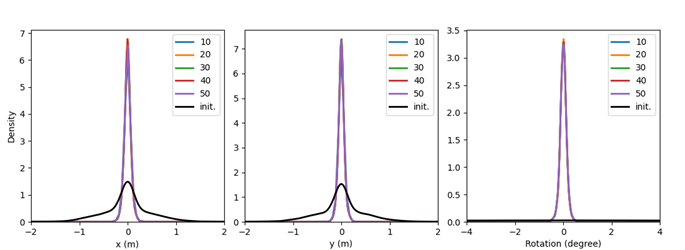}
	\caption{Error distributions after correction with different $N_\text{ransac}$. Init: errors before correction.}
\label{fig:N_ransac}
\end{center}
\end{figure}

\begin{figure}[t]
\begin{center}
		\includegraphics[width=0.75\columnwidth]{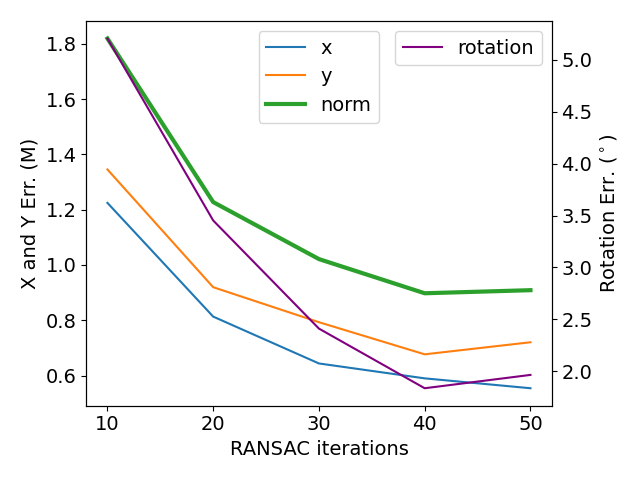}
	\caption{Overall RMSEs after correction with different $N_\text{ransac}$. }
\label{fig:rmse}
\end{center}
\end{figure}
After error correction with our method, the standard deviation of the relative localization is greatly decreased as the example correction result of the initial error of $\sigma = [0.4\SI{}{m}, 0.4\SI{}{m}, 4\SI{}{\degree}]$ in figure \ref{fig:N_ransac} shows. The black lines indicate the initial error distributions, other lines illustrate the error distributions after correction with a different number of RANSAC iterations $N_\text{ransac}$. With all configurations of $N_\text{ransac}$,  the relative errors for most samples can be reduced by a large margin, especially the rotation errors. As shown in the right subplot of figure \ref{fig:N_ransac}, the initial distribution of rotation is nearly flat because of the large standard deviation of the rotation angle in degree. However, after correction, these errors are mostly reduced to less than one degree.

As a quantitative result, figure \ref{fig:rmse} shows the RMSEs of different $N_\text{ransac}$ by averaging over all samples including the outliers. In these outliers, either not enough matchings are found or wrong matchings are used for calculating the final transformation. This will lead to very large errors and therefore result in larger RMSE values. The RMSEs in figure \ref{fig:rmse} that are averaged over all samples and initial error configurations can reflect an overall performance on the dataset. As $N_\text{ransac}$ increases, RMSEs of both translation errors (x, y, and the Euclidean norm of x and y) and rotation errors have a decreasing trend. In the beginning, this trend is stronger than that at the end when $N_\text{ransac}>30$. Therefore, 30 to 40 iterations are the most sufficient configuration of $N_\text{ransac}$ for the proposed method to correct the relative localization errors caused by the global localization errors introduced in the experiments. 

\begin{figure}[th]
\begin{center}
		\includegraphics[width=1\columnwidth]{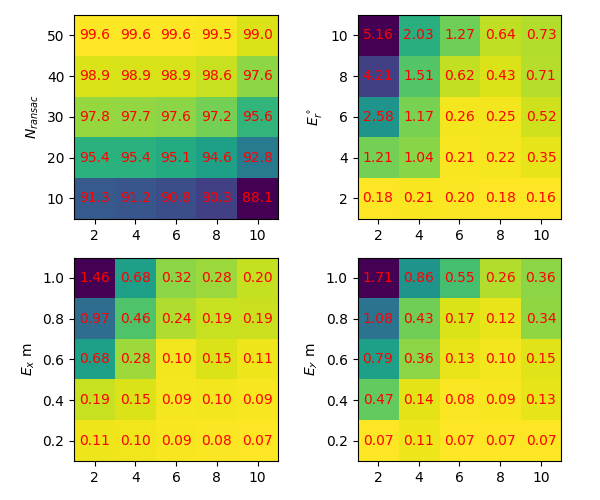}
	\caption{Top left: valid rates with different $N_\text{RANSAC}$ (vertical axis), the other 3 subplots: RMSEs after correction with different initial errors (vertical axis). Horizontal axis represents $thr_\text{cons}$.}
\label{fig:rmse_cons_ransac}
\end{center}
\end{figure}

Different to the overall RMSEs, the top left subplot of figure~\ref{fig:rmse_cons_ransac} shows the valid rates based on different number of iterations (vertical axis) and the thresholds of the number of minimum consensus $thr_\text{cons}$ (horizontal axis). For all $thr_\text{cons}$, the valid rates are increasing as $N_\text{ransac}$ increases and the improvement at smaller $N_\text{ransac}$ is more significant. Besides, the valid rates are, as expected, all decreasing in all $N_\text{ransac}$ configurations as $thr_\text{cons}$ increases. Because larger $thr_\text{cons}$ would lead to  more exclusions of wrong estimation outliers.

\subsection{Performance on different errors}

\begin{figure}[t]
\begin{center}
		\includegraphics[width=0.75\columnwidth]{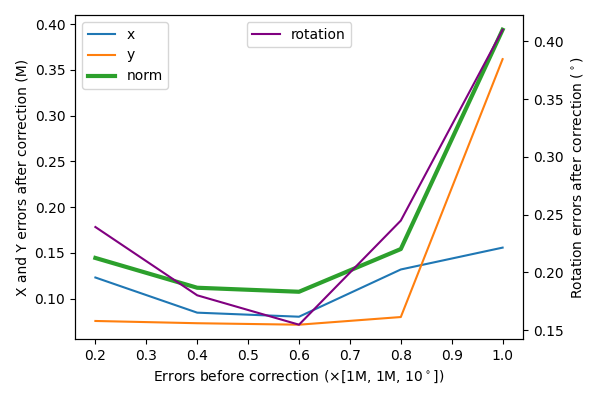}
	\caption{RMSEs with initial different errors, samples filtered by $thr_\text{cons}=10$}
\label{fig:rmse_diff_errors}
\end{center}
\end{figure}

Beside $N_\text{ransac}$, we also plotted RMSEs ($E_x, E_x, E_r$) of different initial localization errors with respect to the consensus thresholds in figure \ref{fig:rmse_cons_ransac}. These plots are all evaluated with $N_\text{ransac}=30$ because it is a sufficient setting regarding the RANSAC iterations and the performance as discussed in the last section. As $thr_\text{cons}$ increases, the RMSEs after correction have mostly a decreasing trend, especially for those with larger initial errors. When the error reduces to a specific level, it reaches its best performance and stops decreasing. For example, the RMSE in the x-axis ($E_x$) decreased from 0.19\SI{}{m} to 0.15\SI{}{m} and then to 0.09\SI{}{m} as the consensus threshold increased from 2 to 6, it then fluctuates in a small range. This fluctuation effect is more apparent in the last row of the RMSE plot of $E_y$ and $E_r$. Besides, as we enlarge the initial standard deviation of the errors, the RMSEs after correction are dropping. This is because our proposed method only searches correspondences in a predefined range. Larger errors will lead to a larger searching range, and therefore also needs more RANSAC iterations. Within limited iterations of searching, the best matching result might not be found and this will, therefore, lead to larger RMSEs. This effect can be better observed with the result of $thr_\text{cons}=10$ in figure \ref{fig:rmse_diff_errors}. At the smaller initial errors, the RMSEs after error correction are all less than 20\SI{}{cm} in the x- and y-axis and the rotation error under 0.25\SI{}{\degree}. The resulting lateral error in y-axis is smaller than the longitudinal error in x-axis because most points of poles and planar objects are regularly aligned along the longitudinal direction which increases the difficulty for matching in this direction. As the error increased to $[1\SI{}{m}, 1\SI{}{m}, 1\SI{}{\degree}]$, the RMSEs are greatly increased---the Euclidean norm of the RMSE of x and y increased from 15\SI{}{cm} to 40\SI{}{cm}, and the RMSE of rotation error from 0.25\SI{}{\degree} to 0.4\SI{}{\degree}.
In this case, a brute-force search in all possible matchings can be used as only a limited number of detected vehicles and poles are in the scenario. However, the benefit of RANSAC in reducing runtime will be lost.

\subsection{Runtime}
\begin{figure}[th]
\begin{center}
		\includegraphics[width=0.47\columnwidth]{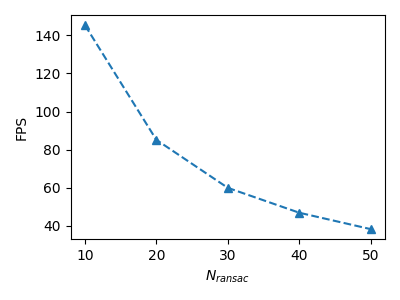}
		\includegraphics[width=0.48\columnwidth]{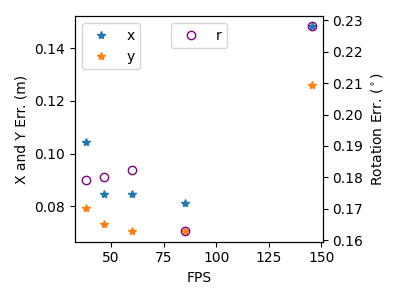}
	\caption{Evaluation on runtimes.}
\label{fig:runtime}
\end{center}
\end{figure}

The number of RANSAC iterations is the most correlated variable to the runtime.  As shown in the left plot of figure \ref{fig:runtime}, the FPS drops as we increase $N_\text{ransac}$. Besides, the dropping slope is becoming more gentle as $N_\text{ransac}$ increases because $N_\text{ransac}$ gradually reaches the number of all possible matches $C^2_{|M^\prime|}$ (see alg. \ref{alg:ransac_err_esti}) in some samples that only have very few anchor points for matching. In the right plot, the FPS with respect to RMSEs of the experiments with initial errors of  $\sigma=[0.4, 0.4m, 4\SI{}{\degree}]$ and $thr_\text{cons}=10$ is shown. At a high FPS rate ($\sim$150), the algorithm generates significantly worse RMSE results than the lower FPS. However, no RMSE reduction is observed in lower FPS. In combination with the previous discussions of valid rate, this reveals that only a small number of RANSAC iterations can already sufficiently find correct matchings between the anchor points of two CAVs, further increasing $N_\text{ransac}$ with the sacrifice of the runtime can only improve the overall performance by increasing the valid rate but not the by reducing the RMSE of all valid estimations which are filtered with $thr_\text{cons}=10$. In other words, if more error estimations are regarded as valid, more hard samples will be included in computing RMSEs, which, therefore, leads to larger RMSEs as shown in the left bottom corner of the right plot in figure \ref{fig:runtime}.

\section{CONCLUSION}\label{sec:conclusion}
In this paper, we proposed an efficient RANSAC-based method to correct the relative localization error between two CAVs that can share the point-cloud-based object detection and semantic segmentation result with each other via CPMs. Our method takes the center points of the detected vehicles and the pole points as anchor points for the RANSAC algorithm to find matches between the observed anchor points of two CAVs and calculate the transformation between the two CAVs based on these matched anchor points. In addition to the anchor points, the points of fences and facades are also used for counting the consensus of RANSAC algorithm and finding the best matches. The result of our experiments on the synthetic dataset COMAP shows that the proposed method can significantly reduce the relative localization errors. The overall performance of the proposed method is related to the initial error, the number of RANSAC iterations, and the consensus threshold for counting the valid error estimations. As far as the anchor points of the two CAVs have enough overlaps and good spatial distribution, the relative localization error can be reduced to less than 20\SI{}{cm}.

Although the proposed method can efficiently and effectively reduce the relative localization error, it has no guarantee to find the best matches because of the randomness of RANSAC. In the worst case, wrong matches can lead to enlarged errors. Therefore, we plan to improve the performance by combining our method, for example, with maximum consensus so that the rotation can be roughly corrected first and therefore the searching range of RANSAC can be reduced. Besides, it is also necessary and meaningful to validate the effectiveness of our method on the real dataset which might have different features on keypoints. For example, the difference of the real traffic pattern might lead to different geometric distributions of the selected keypoints for matching. Moreover, real dataset with more data noise as well as weaker performance quality of object detection and semantic segmentation may result in a lower quality of the detected keypoints. However, it is inauthentic to simulate the patterns and the quality of the detected keypoints, for example, by randomly introducing errors to the keypoints. Instead, it might be more meaningful to test our error correction method on the real dataset and different object detection and semantic segmentation frameworks in the future work.

%\input{acknowledgments}

%=========================
%==== References =========
\renewcommand{\refname}{REFERENCES}
{
	\begin{spacing}{1.17}
		\normalsize
		\bibliography{ref.bib} % Include your own bibliography (*.bib), style is given in isprs.cls
	\end{spacing}
}
\newpage
\section*{APPENDIX}\label{APPENDIX}

\textbf{A: Propagation of the global absolute localization errors of ego vehicle $C_0$ and cooperative vehicle $C_i$ to local relative error.}

Assume the true localization of $C_0$ and $C_1$ is $X_0=(x_0, y_0, \theta_0)$ and $X_i=(x_i, y_i, \theta_i)$, respectively. The localization errors are $dX_0$, $dX_i$. The erroneous localizations are $\tildea{X}_0=X_0+dX_0$, $\tildea{X}_i=X_i+dX_i$. 
\begin{equation*}
    S(m, n) = \begin{bmatrix}
        1 & 0 & m\\
        0 & 1 & n \\
        0 & 0 & 1
    \end{bmatrix},\quad
    R(l) = \begin{bmatrix}
    cos(l) & -sin(l) & 0\\
    sin(l) & cos(l) & 0 \\
    0 & 0 & 1
    \end{bmatrix}
\end{equation*}
%\begin{equation*}
\begin{align*}
  PC^{i,0} 
  &=R(-\theta_0)S(-x_0,-y_0)\cdot PC^{i,g}\\
  &=R(-\theta_0)S(-x_0,-y_0)S(x_i,y_i)R(\theta_i)\cdot PC^{i,i}\\
  &= \Delta T \cdot \tildea{PC}^{i,0}
   = \Delta T \cdot R(-\tildea{\theta_0})S(-\tildea{x_0}, -\tildea{y_0})\cdot \tildea{PC}^{i,g}\\
  &= \Delta T \cdot R(-\tildea{\theta_0})S(-\tildea{x_0}, -\tildea{y_0})S(\tildea{x_i}, \tildea{y_i})R(\tildea{\theta_i})\cdot PC^{i,i}\\
\end{align*}
\begin{align*}
    \Longrightarrow & R(-\theta_0)S(-x_0,-y_0)S(x_i,y_i)R(\theta_i)\\
    & =\Delta T \cdot R(-\tildea{\theta_0})S(-\tildea{x_0}, -\tildea{y_0})S(\tildea{x_i}, \tildea{y_i})R(\tildea{\theta_i})\\
    \Longrightarrow
    & \Delta T =  R(-\theta_0)S(-x_0,-y_0)S(x_i,y_i)R(\theta_i)\cdot \\ 
    & R^T(\tildea{\theta_i})S^T(\tildea{x_i}, \tildea{y_i})S^T(-\tildea{x_0}, -\tildea{y_0})R^T(-\tildea{\theta_0})
\end{align*}
%\end{equation*}
\end{document}